\pdfoutput=1

\documentclass[11pt]{article}

\usepackage[final]{acl}

\usepackage{times}
\usepackage{latexsym}
\usepackage{amssymb}
\usepackage{adjustbox}
\usepackage{amsmath}
\usepackage[T1]{fontenc}

\usepackage[utf8]{inputenc}

\usepackage{microtype}

\usepackage{inconsolata}

\usepackage{graphicx}
\usepackage[capitalize]{cleveref}
\usepackage{enumitem,kantlipsum}
\usepackage{booktabs}
\usepackage{multirow}
\usepackage{tabularx}
\usepackage{xcolor}
\usepackage{colortbl}
\usepackage{makecell}

\renewcommand{\b}[1]{\textbf{#1}}

\newcommand{\shortname}{\texttt{SpanDec}}
\newcommand{\efficientmodel}{\texttt{SF-SpanDec}}

\usepackage{color}
\usepackage{xcolor}
\usepackage{xspace}
\usepackage{bigstrut}

\Crefname{equation}{Eq.}{Eqs.}
\Crefname{appendix}{App.}{Apps.}
\Crefname{figure}{Fig.}{Figs.}
\Crefname{table}{Tab.}{Tabs.}
\Crefname{section}{Sec.}{Secs.}
\Crefname{algorithm}{Alg.}{Algs.}

\newcommand{\good}[1]{\textcolor{green!55!black}{\textbf{#1}}}
\newcommand{\midq}[1]{\textcolor{orange!85!black}{\textbf{#1}}}
\newcommand{\bad}[1]{\textcolor{red!70!black}{\textbf{#1}}}

\newcommand{\Hi}{\good{High}}
\newcommand{\Md}{\midq{Med}}
\newcommand{\Lo}{\bad{Low}}

\usepackage[table]{xcolor}
\usepackage{array}

\definecolor{softgreen}{HTML}{DDEED7}
\definecolor{softyellow}{HTML}{FFF4CC}
\definecolor{softorange}{HTML}{FBE5D6}
\definecolor{softred}{HTML}{F8D7DA}

\newcolumntype{G}{!{\color{white}\vrule width 3pt}}

\title{Decoding Text Spans for Efficient and Accurate Named-Entity Recognition}

\author{Andrea Maracani, Savas Ozkan, Junyi Zhu, Sinan Mutlu, Mete Ozay \\
         Samsung Research UK, Staines-upon-Thames, United Kingdom \\ \texttt{\{a.maracani, savas.ozkan, junyi.zhu, s.mutlu, m.ozay\}@samsung.com} }

\begin{document}
\maketitle

\begin{abstract}
Named Entity Recognition (NER) is a key component in industrial information extraction pipelines, where systems must satisfy strict latency and throughput constraints in addition to strong accuracy. State-of-the-art NER accuracy is often achieved by span-based frameworks, which construct span representations from token encodings and classify candidate spans. However, many span-based methods enumerate large numbers of candidates and process each candidate with marker-augmented inputs, substantially increasing inference cost and limiting scalability in large-scale deployments. 
In this work, we propose \shortname{}, an efficient span-based NER framework that targets this bottleneck. Our main insight is that span representation interactions can be computed effectively at the final transformer stage, avoiding redundant computation in earlier layers via a lightweight decoder dedicated to span representations. We further introduce a span filtering mechanism during enumeration to prune unlikely candidates before expensive processing. Across multiple benchmarks, \shortname{} matches competitive span-based baselines while improving throughput and reducing computational cost, yielding a better accuracy--efficiency trade-off suitable for high-volume serving and on-device applications.
\end{abstract}

\section{Introduction}

Named Entity Recognition (NER) is a core component of industrial information extraction pipelines, enabling applications such as clinical and biomedical coding, legal document parsing, and high-volume email understanding for reminder and task generation~\citep{info10080248,Dozier2010,10.5120/ijca2020920526}. In production settings, NER models are often deployed as always-on services and must satisfy strict latency and throughput targets under cost constraints (e.g., GPU/CPU budget and total cost of ownership), making \emph{accuracy--efficiency trade-offs} at least as important as peak F1/accuracy~\citep{wang-etal-2020-minilm}.

\begin{figure}[t]
\centering
\includegraphics[width=0.48\textwidth]{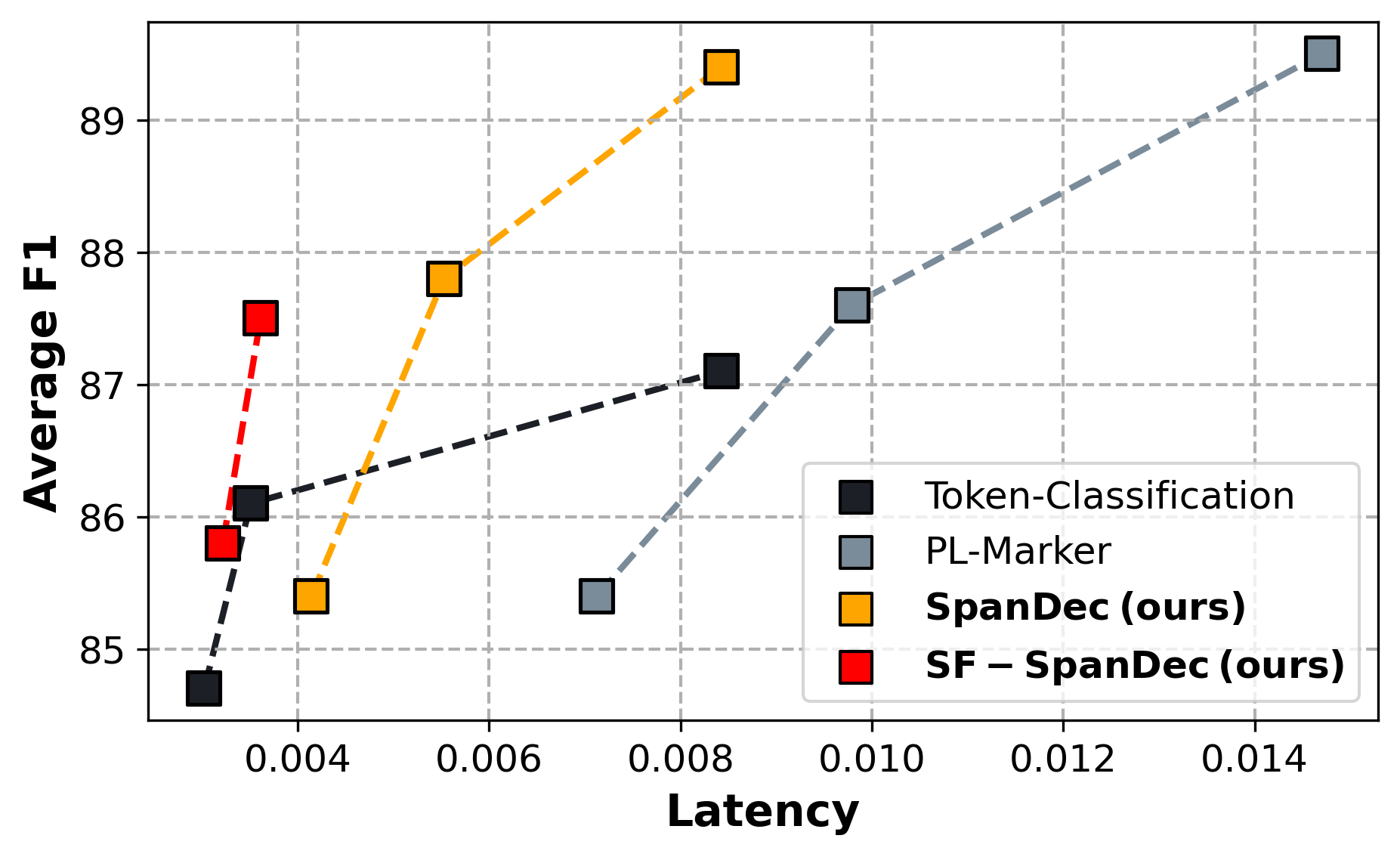}
\vspace{-8mm}
\caption{\b{F1 scores vs. Latency}. Our methods achieve a better accuracy--efficiency trade-off under realistic serving constraints. Same color markers represent encoder models of different sizes (Details in Sec.~\ref{sec:exp}).}
\label{fig:plot}
\end{figure}

A widely used baseline is token classification with pre-trained transformer encoders, which performs per-token labeling with a lightweight prediction head and offers strong accuracy after fine-tuning~\citep{devlin-etal-2019-bert}. While effective, token-level labeling can struggle to represent complete entity spans for ambiguous boundaries (and is less naturally compatible with span-level reasoning), motivating span-based formulations that explicitly construct and classify candidate spans~\citep{zhong-chen-2021-frustratingly,wang-etal-2021-automated}. Among span-based approaches, \citet{ye-etal-2022-packed} propose \emph{Packed Levitated Markers} (PL-Marker), which inserts marker token pairs to focus the encoder on span representations and has become a strong and widely-adopted backbone for span-based NER~\citep{ye-etal-2022-packed}.

Despite their accuracy benefits, marker-based span methods introduce non-trivial serving overhead. In contrast to token classification, which does not require enumerating and processing large numbers of candidate spans, span-based methods can consider $O(L^2)$ candidates, and even variants that cap the maximum span length may still incur substantial extra computation due to repeated processing of marker-augmented inputs (see \cref{sec:bg}). In large-scale deployments, this overhead can become the dominant bottleneck and can force practitioners to reduce batch sizes, shorten context windows, or fall back to simpler models.

Recent LLM-based approaches also tackle NER via prompting or text generation, sometimes offering strong generalization and low-label adaptability; however, their autoregressive decoding and large parameter footprints typically yield prohibitively high latency and cost for high-throughput serving~\citep{wang-etal-2025-gpt,wang-etal-2023-instructuie}. 

\begin{figure}[t]
\centering
\includegraphics[width=0.48\textwidth]{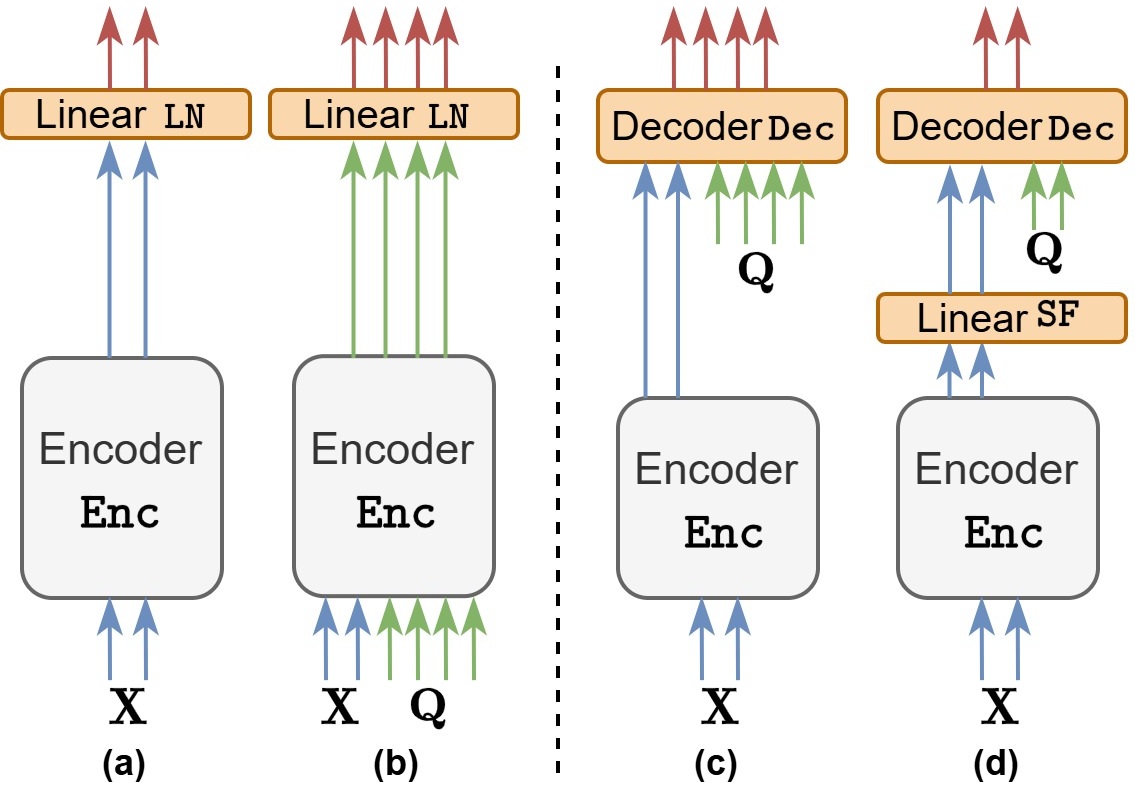}
\vspace{-6mm}
\caption{\b{Diagram of Model Flows.} \b{(a)} Token classification approach, \b{(b)} span classification approach, \b{(c)} \shortname{} (ours), and \b{(d)} \efficientmodel{} (ours).}
\label{fig:architecture}
\end{figure}

In this work, we target the practical bottleneck of PL-Marker-style NER and propose two complementary optimizations that preserve the modeling benefits of span representations while substantially reducing inference cost:

\vspace{-1mm}
\begin{itemize}[leftmargin=*]
\item \textbf{Decoupled span processing (\shortname{}).}
To reduce redundant computation, we decouple span-marker processing from the main encoder. Concretely, we replace the final encoder layer used by PL-Marker with a lightweight decoder dedicated to span representations (see \cref{fig:architecture}c). The text tokens are encoded once as usual, while span-marker interactions are handled only in the decoder, avoiding repeated processing of marker tokens in earlier layers. We keep the overall parameter budget comparable to the original model by trading an encoder layer for the decoder. \shortname{} delivers significant throughput gains without degrading accuracy (Sec.~\ref{sec:method}).
\vspace{-1mm}
\item \textbf{Early span filtering (\efficientmodel{}).}
We further improve end-to-end efficiency with a simple and effective span filtering mechanism: we train a lightweight binary classifier to predict tokens that contain no entity mentions (see \cref{fig:architecture}d). At inference time, \efficientmodel{} prunes the candidate set aggressively, often retaining only a small fraction of spans (e.g., $\sim$15\%),so the decoder processes only likely entity spans (Sec.~\ref{sec:sf}).
\end{itemize}

Across multiple NER benchmarks, our approach matches or exceeds the accuracy of PL-Marker while increasing throughput by up to $2.7\times$ and reducing computation (GFLOPs) by up to $8.2\times$. Moreover, compared to standard token classification, our span-centric approach achieves higher average F1 (+1.8\%) while remaining viable under high-throughput serving constraints.

\section{Related Work}

\begin{table*}[t!]
\centering
\footnotesize
\setlength{\tabcolsep}{5pt}
\renewcommand{\arraystretch}{1.12}

\begin{tabularx}{\textwidth}{@{}p{2.35cm} X c c c p{4.3cm}@{}}
\toprule
\textbf{Paradigm} &
\textbf{Method example} &
\textbf{Acc. $\uparrow$} &
\textbf{Latency $\downarrow$} &
\textbf{Scalability $\uparrow$} &
\textbf{Typical deployment issue} \\
\midrule

Token classification &
BERT tagging \citeyearpar{devlin-etal-2019-bert} &
\Md &
\good{Low} &
\good{High} &
Boundary / nesting limitations; weaker span modeling \\

\addlinespace[2pt]

Span enumeration &
DyGIE++ \citeyearpar{wadden-etal-2019-dygiepp} &
\Hi &
\bad{High} &
\Lo &
Quadratic candidates; heavy compute/memory for long inputs \\

\addlinespace[2pt]

Marker-based span (encoder) &
PL-Marker \citeyearpar{ye-etal-2022-packed} &
\Hi &
\bad{High} &
\midq{Med--Low} &
Markers processed in all encoder layers; latency bottleneck \\

\addlinespace[2pt]

Generative, prompted (LLM) &
InstructUIE \citeyearpar{wang-etal-2023-instructuie} &
\midq{Med--High} &
\bad{V.High} &
\Lo &
Autoregressive decoding cost; controllability / hallucination; cost \\

\addlinespace[2pt]

\midrule

\textbf{Decoupled span decoding (ours)} &
\shortname{}, \efficientmodel{} &
\Hi &
\good{Low} &
\good{High} &
Extra module integration; thresholding trade-off \\

\bottomrule
\end{tabularx}

\caption{Industrial view of NER paradigms: qualitative accuracy--latency--scalability trade-offs. Arrows indicate the preferred direction for deployment.}
\label{tab:ner_paradigms}
\end{table*}

Table~\ref{tab:ner_paradigms} summarizes common NER paradigms through an industrial lens, highlighting that span- and marker-based methods can improve boundary modeling at the cost of latency and scalability, while LLM-based prompting offers flexibility but remains expensive for high-throughput deployments.

\paragraph{Token classification for NER.}
The dominant practical baseline for NER is token-level sequence labeling with a transformer encoder and a lightweight classification head~\citep{devlin-etal-2019-bert}. Token classification is appealing in production because it scales linearly with input length and maps cleanly to efficient batching and serving. Recent work also explores task-specialized encoders trained or adapted for NER, including pre-training with large-scale (possibly LLM-assisted) NER-style signals to improve data efficiency~\citep{bogdanov-etal-2024-nuner}.

\paragraph{Span-based and marker-based NER.}
Span-based NER replaces per-token labeling with explicit span enumeration and span classification, which can better capture whole-entity boundaries and support joint reasoning across spans~\citep{zhong-chen-2021-frustratingly,wang-etal-2021-automated}. Classic span-centric IE frameworks such as DyGIE++ enumerate candidate spans and iteratively refine span representations using contextual signals~\citep{wadden-etal-2019-dygiepp}. PL-Marker strategically inserts levitated markers to obtain stronger span representations from pre-trained encoders and has become a strong backbone for span-style NER~\citep{ye-etal-2022-packed}. While effective, these approaches can incur substantial computation due to (i) the number of candidate spans and (ii) repeated processing of marker-augmented inputs.

\paragraph{Improving efficiency in span-based extraction.}
Efficiency constraints have motivated both architectural and algorithmic changes. A common direction is to reduce the cost of the underlying encoder via compression techniques such as knowledge distillation (e.g., MiniLM) to improve latency without severe accuracy loss~\citep{wang-etal-2020-minilm}. Another direction is to reduce the number of candidates processed: span filtering or staged detection/classification can prune the search space before expensive classification. For example, Split-NER decomposes NER into a two-stage pipeline (span detection $\rightarrow$ span classification), casting both stages as Question Answering (QA) and training them as separate components~\citep{arora-park-2023-split-ner}. Our work is complementary but different in nature: rather than reformulating NER into a multi-stage QA pipeline, we keep the PL-Marker formulation intact and reduce its serving-time overhead by (1) avoiding redundant marker computation in the encoder via a lightweight span decoder and (2) pruning spans early so the decoder processes only likely candidates.


\paragraph{LLMs and instruction-based extraction.}
Recent trends use LLMs and prompting/instruction tuning for information extraction, either directly (e.g., generative or prompted NER) or to improve supervision. GPT-NER adapts large language models to NER via prompting and verification strategies and reports competitive results in low-resource settings~\citep{wang-etal-2025-gpt}. InstructUIE proposes instruction-tuning to unify multiple IE tasks in a text-to-text framework~\citep{wang-etal-2023-instructuie}. In parallel, compact “generalist” NER models such as GLiNER aim to support flexible schemas while keeping inference parallel and efficient compared to autoregressive generation~\citep{zaratiana-etal-2024-gliner}. Finally, verification-style post-processing with LLM reasoning has been explored to improve faithfulness in domain NER (e.g., biomedical) by revising model outputs using external knowledge~\citep{kim-etal-2024-verifiner}. Compared to these lines of work, we focus on high-throughput deployment of span-based NER under tight latency constraints, providing a practical accuracy--efficiency improvement to PL-Marker-style systems.

\section{Background and Preliminaries of NER}
\label{sec:bg}
Let ${\mathbf{X} = [\mathbf{x}_1, \mathbf{x}_2, \ldots, \mathbf{x}_L]}$ represent an input sequence of $L$ tokens, such as a sentence or document.
Let $\mathcal{E} = \{e_0, e_1, \ldots, e_M\}$ is a pre-defined set of $M$ entity types with an additional $e_0 = \texttt{``O"}$ tag, denoting the ``outside'' class (i.e., absence of any named entity).
The objective of NER is twofold:

\vspace{+0.3em}
\noindent 1. To identify the contiguous spans of tokens, with each span denoted by $\mathbf{x}_{[i,j]} = [\mathbf{x}_i, \ldots, \mathbf{x}_j]$, composing a single named entity.

\vspace{+0.3em}
\noindent 2. To classify each identified $\mathbf{x}_{[i,j]}$ with one of the specific entity types $e_k \in \mathcal{E}$.

\paragraph{Token Classification.} 
A prevalent approach to NER frames the task as a sequence labeling problem.
Each token $\mathbf{x}_i$ in the input sequence $\mathbf{X}$ is first mapped to a contextualized vector representation $\mathbf{z}_i \in \mathbb{R}^d$ using a text encoder, $\texttt{Enc}(\cdot)$ by
\[
[\mathbf{z}_1, \mathbf{z}_2, \ldots, \mathbf{z}_L] = \texttt{Enc}([\mathbf{x}_1, \mathbf{x}_2, \ldots, \mathbf{x}_L]).
\]
In this approach, to effectively describe entity boundaries and distinguish between adjacent entities of the same type, tagging schemes such as BIO (Beginning, Inside, Outside) are commonly employed.
In this scheme, the set of entity types $\mathcal{E}$ is expanded to $\mathcal{E}_{\texttt{BI}}$ that represents entity tags. Intuitively, for each entity type $e_k \in \mathcal{E} \setminus \{e_0\}$, two tags are created: \texttt{B-$e_k$} (Beginning of entity $e_k$) and \texttt{I-$e_k$} (Inside of entity $e_k$). The \texttt{O} tag remains for tokens outside any entity.

Subsequently, a classifier, typically a linear layer, $\texttt{LN}(\cdot)$ is applied to predict an entity tag $e^\texttt{BI}_k \in \mathcal{E}_{\texttt{BI}}$ for each $\mathbf{z}_i$.

\paragraph{Span Classification.}
An alternative framework for NER is the span classification. In this formulation, all possible contiguous subsequences of tokens $\mathbf{x}_{[i,j]}$ from the input sequence $\mathbf{X}$ are considered candidate named entities, and we refer to the ranges $[i,j]$ as \textit{spans}.
Each candidate span $[i,j]$ is represented as a vector $\mathbf{q}_{ij}, \forall i, j$, often derived from the contextualized embeddings of its constituent tokens (e.g., by concatenating the representations of its start and end tokens, or by a pooling operation over the span's tokens). 

Span representations ${\mathbf{Q}=[\mathbf{q}_{00}, \mathbf{q}_{01},\ldots,\mathbf{q}_{LL}]}$ and the input sequence $\mathbf{X}$ are fed into the encoder $\texttt{Enc}(\cdot)$ and linear layer $\texttt{LN}(\cdot)$ to predict corresponding entity types from $\mathcal{E}$, including the \texttt{O} type if it is not a valid entity.
Span classification approaches consistently  yield improved performance compared to token classification techniques, particularly for nested or complex text entity structures. However, they can also introduce greater computational complexity due to the enumeration of spans, which can be quadratic $O(L^2)$ for the exhaustive enumeration and/or can be linear $O(L)$ when the maximum length of span is fixed ($L$ is the sequence length).

\section{Our Method}
\label{sec:method}
Our proposed method builds upon the span classification framework, offering a novel method designed for simplicity and computational efficiency relative to existing span-based approaches.

\paragraph{Span Representation.}
We represent each candidate span $[i, j]$, with $1 \leq i \leq j \leq L$, using two special learnable vectors: a \textit{start} marker $\mathbf{m}_{\text{start}}$ and an \textit{end} marker $\mathbf{m}_{\text{end}}$. 
These markers are augmented with positional embedding of the corresponding input text tokens to encode the span boundaries by $\mathbf{q}_i = \mathbf{m}_{\text{start}} + \text{pos}(i)\,\,$ and $\quad \mathbf{q}_j = \mathbf{m}_{\text{end}} + \text{pos}(j)$
where $\text{pos}(\cdot)$ denotes a positional encoding function. The pair $\mathbf{q}_{ij}=(\mathbf{q}_i, \mathbf{q}_j)$ serves as the query representation for the span of candidate entity $\mathbf{x}_{[i,j]}$.

\paragraph{Span Decoder.}
To classify the span, we propose a lightweight decoder $\texttt{Dec}(\cdot)$ composed of a pre-norm cross-attention layer $\texttt{Att}(\cdot)$ followed by a multilayer perceptron $\texttt{MLP}(\cdot)$ and a classification layer $\texttt{LN}(\cdot)$. Each span pair $\mathbf{q}_{ij}=(\mathbf{q}_i, \mathbf{q}_j)$ attends to the encoder outputs $[\mathbf{z}_1, \ldots, \mathbf{z}_L]$ as well as to each others (but not to other span pairs). Specifically, for the span $\mathbf{x}_{[i,j]}$, the input key $\mathbf{K}_{ij}$ and value $\mathbf{V}_{ij}$ sequences in the cross-attention are defined by:
\[
    \mathbf{K}_{ij} = \mathbf{V}_{ij} = [\mathbf{q}_i, \mathbf{q}_j, \mathbf{z}_1, \ldots, \mathbf{z}_L].
\]
This formulation allows the span representation to incorporate contextual information from the full input sequence while also enhancing the focus between the span boundaries. The two outputs of the cross-attention layer $\texttt{Att}$ (denoted by $\mathbf{q}_i'$ and  $\mathbf{q}_j'$) are passed through a shared $\texttt{MLP}$, concatenated and classified using a linear layer $\texttt{LN}$. As an output, a vector $\hat{\mathbf{y}}_{ij}$ containing logits of the classes of the entity types $\mathcal{E}$ is calculated by
\[
    \hat{\mathbf{y}}_{ij} = \texttt{LN}([\texttt{MLP}(\mathbf{q}_i'); \texttt{MLP}(\mathbf{q}_j')]),
\]
where $[\cdot;\cdot]$ denotes the concatenation operation in the channel dimension.

\paragraph{Computational Efficiency.}
Compared to vanilla span-based NER approaches such as PL-Marker, which appends span markers to the input of the model and propagate them through all layers of the encoder, our approach is more computationally efficient. Precisely, span representations are introduced only at the decoding stage in our method. Thereby, our method significantly reduces the computational overhead, especially when processing a large number of spans (e.g., longer input text).

\subsection{Efficient Span Selection}
\label{sec:sf}

To further improve the computational efficiency of our method while maintaining high performance, we introduce a lightweight Span Filtering ($\texttt{SF}$) mechanism based on binary token-level classification. Formally, before executing the span decoder $\texttt{Dec}$, we apply a linear classifier to contextualized vector representations $[\mathbf{z}_1, \ldots, \mathbf{z}_L]$ produced by the encoder $\texttt{Enc}$. The auxiliary $\texttt{SF}$ classifier  is trained to predict whether a token belongs to the \texttt{O} class (i.e., not part of any named entity) or to any entity class with the information estimated only from the encoder. 

Importantly, the weights of the \texttt{SF} classifier and of the span decoder \texttt{Dec} are optimized jointly during the training phase. At inference time, we use the output of \texttt{SF} classifier to identify tokens that are likely not part of an entity by applying a confidence threshold. Spans that enclose tokens predicted as \texttt{O} are discarded and not passed to the decoder. As a result, the number of candidate spans processed by the span decoder is significantly reduced.

This filtering step introduces negligible overhead, but yields substantial gains in efficiency, as it limits the unnecessary computations in the decoder step with a much smaller set of potentially valid spans. Empirically, we observe that this barely causes negative impact on the overall performance while dramatically reducing the number of spans fed to the decoder, leading to a promising efficiency-accuracy trade-off. 

\section{Implementation and Setup}
\label{sec:implement}
We conducted our experiments using the PyTorch framework and leveraged pre-trained models available through the Hugging Face Transformers library~\cite{wolf-etal-2020-transformers}. We implemented our proposed methodologies (\shortname{} and \efficientmodel{}) and the token-classification baseline, from scratch, using PyTorch Lightning. For comparison with PL-Marker, we utilized the SpanMarker~\cite{Aarsen_SpanMarker} library, which is built upon Hugging Face components and designed for efficient performance. All experiments were performed on a cluster of 8 NVIDIA A40 GPUs. For testing and throughput evaluation, a single NVIDIA A40 GPU was used. Mixed-precision training was always employed.

\paragraph{Models.} We employ and test three widely adopted text encoders of varying sizes: \textit{MiniLM} (33M), \textit{BERT-Base} (110M), and \textit{RoBERTa-Large} (355M). For \shortname{} , we remove the final layer of each encoder to maintain a comparable overall parameter size after integrating the decoder. With the goal of improving the computational efficiency for on-device applications,
we study our Span Filtering model (\efficientmodel{}) with lightweight encoders. Accordingly, we conduct experiments using MiniLM and also include results for BERT-Base to offer a broader comparison.

\paragraph{Datasets.} We select four widely used NER benchmarks: (1) \textit{CoNLL++}~\cite{wang2019crossweigh}, a corrected version of the standard CoNLL03 dataset~\cite{tjong-kim-sang-de-meulder-2003-introduction}; (2) \textit{CrossNER}~\cite{liu2020crossner}, a cross-domain dataset covering five diverse domains; (3) \textit{OntoNotes v5} (English)~\cite{hovy-etal-2006-ontonotes}, a large-scale dataset comprising multiple genres of text; and (4) \textit{BC5CDR}~\cite{DBLP:journals/biodb/LiSJSWLDMWL16}, which contains PubMed articles annotated with chemical and disease entities (see Tab.~\ref{tab:datasets}).

\paragraph{Metrics.} Following prior work, we report the F1 score (using the evaluation techniques provided by the seqeval library~\cite{seqeval}), the GFLOPs (see \cref{sec:flops}) and the Throughput (samples processed per second) of different methodologies.

\begin{table}[t!]
    \centering
    \centering
    \begin{adjustbox}{max width=\columnwidth}
    \begin{tabular}{lcccc}
    
    \toprule
    & \textbf{CoNLL++} & \textbf{CrossNER} & \textbf{OntoNotes5} & \textbf{BC5CDR} \\
    \cmidrule(lr){2-5}
    \textbf{Entity Classes} & 5 & 40 & 19 & 3 \\
    \cmidrule(lr){1-5}
    \textbf{Train Samples} & 4.5k & 20k & 28k & 11k \\
    \textbf{Train Words}   & 272k & 364k & 1547k & 218k \\ 
    \cmidrule(lr){1-5}
    \textbf{Test Samples} & 817 & 2.5k & 3.2k & 5.9k \\
    \textbf{Test Words}   & 50k & 96k & 179k & 116k \\ 
    
    \bottomrule
    \end{tabular}
    \end{adjustbox}

    \caption{\b{Datasets.} Information about the datasets used in our experiments.}
    \label{tab:datasets}
\end{table}

\paragraph{Training Procedure.} Standard hyperparameters and training schedules were used for all models. MiniLM was trained for 100 epochs on each dataset with a learning rate of $5 \times 10^{-5}$. BERT-B and RoBERTa-L were trained for 25 epochs on all datasets, except for the CrossNER dataset, where 100 epochs were used. The learning rate of newly initialized weights (decoder and final linear layer) was multiplied by 10, a common practice. A OneCycle learning rate schedule with a warm-up ratio of 0.03 was applied, and a global batch size of 64 was used. Gradient clipping was set to 1.0. Following the SpanMarker implementation, we assumed a maximum entity length of 8 words. Entities exceeding this length were treated as errors, although there is just a negligible number of such cases in the datasets considered. Model weights were optimized using the AdamW optimizer with a weight decay of 0.01. 



\section{Results}

\begin{table*}[h!]
    \centering
    \vspace{-2em}
    \begin{adjustbox}{max width=\textwidth}
    \begin{tabular}{llc!{\color{white}\vrule width 3pt}cccccc}
    \toprule
    \textbf{Model} & \textbf{Strategy} & \textbf{Throughput} $\uparrow$ & \textbf{GFLOPs} $\downarrow$ & \textbf{CoNLL++} & \textbf{CrossNER} & \textbf{OntoNotes5} & \textbf{BC5CDR} & \textbf{Avg.} \\
    \cmidrule(lr){1-2}\cmidrule(l{1pt}r{1pt}){3-3}\cmidrule(l{1pt}r{1pt}){4-4}\cmidrule(lr){5-8}\cmidrule(lr){9-9}
    \multirow{3}{*}{MiniLM}
        & Token Classif.      & \cellcolor{softgreen}\b{330.8} \small(1$\times$)   & \cellcolor{softgreen}\b{1.9} \small(1$\times$)   & 92.9 & 71.2 & 87.8 & 87.1 & 84.7 \\
        & PL-Marker           & \cellcolor{softorange}140.3 \small(0.42$\times$)   & \cellcolor{softorange}15.8 \small(8.3$\times$)   & 91.7 & 73.9 & 87.9 & \b{88.0} & 85.4 \\
        & \b{SpanDec (ours)}  & \cellcolor{softyellow}241.3 \small(0.73$\times$)   & \cellcolor{softyellow}2.9 \small(1.5$\times$)    & \b{93.7} & \b{74.6} & \b{88.4} & \b{88.0} & \b{86.2} \\
         
    \cmidrule(lr){1-2}\cmidrule(l{1pt}r{1pt}){3-3}\cmidrule(l{1pt}r{1pt}){4-4}\cmidrule(lr){5-8}\cmidrule(lr){9-9}
    
    \multirow{3}{*}{BERT-B}
        & Token Classif.      & \cellcolor{softgreen}\b{299.1} \small(1$\times$)   & \cellcolor{softgreen}\b{7.5} \small(1$\times$)   & 92.9 & 74.5 & 89.5 & 87.4 & 86.1 \\
        & PL-Marker           & \cellcolor{softorange}102.1 \small(0.34$\times$)   & \cellcolor{softorange}62.5 \small(8.3$\times$)   & 93.2 & 78.5 & 90.0 & \b{88.5} & 87.6 \\
        & \b{SpanDec (ours)}  & \cellcolor{softyellow}180.9 \small(0.61$\times$)   & \cellcolor{softyellow}11.5 \small(1.5$\times$)   & \b{93.6} & \b{78.7} & \b{90.3} & 88.4 & \b{87.8} \\
    
    \cmidrule(lr){1-2}\cmidrule(l{1pt}r{1pt}){3-3}\cmidrule(l{1pt}r{1pt}){4-4}\cmidrule(lr){5-8}\cmidrule(lr){9-9}
    
    \multirow{3}{*}{RoBERTa-L}
        & Token Classif.      & \cellcolor{softgreen}\b{187.0} \small(1$\times$)   & \cellcolor{softgreen}\b{26.8} \small(1$\times$)  & 94.3 & 76.1 & 89.9 & 88.2 & 87.1 \\
        & PL-Marker           & \cellcolor{softorange}68.0 \small(0.36$\times$)    & \cellcolor{softorange}222.7 \small(8.3$\times$)  & \b{95.4} & \b{80.2} & \b{91.4} & \b{90.9} & \b{89.5} \\
        & \b{SpanDec (ours)}  & \cellcolor{softyellow}118.7 \small(0.64$\times$)   & \cellcolor{softyellow}33.9 \small(1.3$\times$)   & \b{95.4} & 79.9 & 91.3 & 90.8 & 89.4 \\
    
    \bottomrule
    \end{tabular}
    \end{adjustbox}
    \vspace{-0.5em}
    \caption{\b{Results for SpanDec.} F1 scores on CoNLL++, CrossNER, OntoNotes5 and BC5CDR benchmarks using different encoder models. The best performance per dataset and encoder model is highlighted in \textbf{bold}. The \textit{Throughput} shows the samples processed per second by models (relative value compared to token classification reported in parenthesis). For the \textit{GFLOPs}, we take the input length as 44, which is the average value of all datasets.}
    \label{tab:main_table_v2}
\end{table*}

\begin{table}
    \centering
    \begin{adjustbox}{max width=\columnwidth}
    \begin{tabular}{lccccc}
    \toprule
    \textbf{Method} & \textbf{Params} & \textbf{Throughput} $\uparrow$ & \textbf{BC5} & \textbf{CoNLL} & \textbf{O-Notes} \\

    \cmidrule(lr){1-1}\cmidrule(lr){2-2}\cmidrule(l{1pt}r{1pt}){3-3}\cmidrule(lr){4-6}
    LLaMA2-7B    & 7B    & \cellcolor{softred}{$< 9$ \small ($0.08\times$)} & - & 76.53 & 52.78 \\
    LLaMA2-13B   & 13B   & \cellcolor{softred}{$< 7$ \small ($0.06\times$)} & - & 68.24 & 38.32 \\
    LLaMA3-8B    & 8B    & \cellcolor{softred}{$< 9$ \small ($0.08\times$)} & - & 85.37 & 22.34 \\
    SOLAR        & 10.7B & \cellcolor{softred}{$< 8$ \small ($0.07\times$)} & - & 76.03 & 64.84 \\
    Mistral-12B  & 12B   & \cellcolor{softred}{$< 7$ \small ($0.06\times$)} & - & 74.23 & 59.30 \\

    \cmidrule(lr){1-1}\cmidrule(lr){2-2}\cmidrule(l{1pt}r{1pt}){3-3}\cmidrule(lr){4-6}
    SplitNER \citeyearpar{arora-park-2023-split-ner}         & 0.35B & \cellcolor{softyellow}104.1 \small ($0.88\times$) & -    & -    & 90.9 \\
    InstructUIE \citeyearpar{wang-etal-2023-instructuie}     & 13B   & \cellcolor{softred}{$< 5.8$ \small ($0.05\times$)} & 89.0 & 91.5 & 88.6 \\
    UniNER \citeyearpar{zhou2023universalner}                & 7B    & \cellcolor{softred}{$< 6.4$ \small ($0.05\times$)} & 89.0 & 93.3 & 89.9 \\
    GLiNER \citeyearpar{zaratiana-etal-2024-gliner}          & 0.35B & \cellcolor{softorange}76.2 \small ($0.64\times$)   & 88.7 & 92.5 & 88.1 \\
    GPT-NER \citeyearpar{wang-etal-2025-gpt}                 & 175B  & \cellcolor{softred}{$< 5$ \small ($0.04\times$)}    & -    & 90.9 & 82.2 \\

    \cmidrule(lr){1-1}\cmidrule(lr){2-2}\cmidrule(l{1pt}r{1pt}){3-3}\cmidrule(lr){4-6}
    \textbf{SpanDec} (RoBERTa-L) & 0.35B & \cellcolor{softgreen}\textbf{118.7} \small ($1\times$) & \b{90.8} & \b{95.4} & \b{91.3} \\

    \bottomrule
    \end{tabular}
    \end{adjustbox}
    \caption{\b{Comparison with SOTA.} Top section: results for 1-shot generative inference with open LLMs from \cite{zhu2024open}. Mid section: results of SOTA.}
    \label{tab:sota}
\end{table}

\label{sec:exp}


\paragraph{Decoding Spans is All You Need.}
As shown in Tab~\ref{tab:main_table_v2}, our method outperforms PL-Marker in terms of computational efficiency and achieves similar F1 score. For instance, with MiniLM and compared to the token-classification baseline, PL-Marker provides 42\% of the throughput and 0.7\% average accuracy increase, whereas our method provides a notable 73\% of the throughput with an average performance boost of 1.8\%.
A similar trend is observed with BERT-Base: our approach not only improves performance, but also maintains better efficiency compared to PL-Marker. Even with RoBERTa-Large, our method achieves competitive performance comparable to PL-Marker while being notably more efficient during inference.

These results support the superiority of our design, demonstrating that a lightweight decoder alone is sufficient for processing and classifying spans both effectively and efficiently.

\noindent \textbf{Comparison with SOTA.} In Tab.~\ref{tab:sota} we compare \shortname{} (RoBERTa-L encoder) with other SOTA models in terms of model parameters, throughput and F1 score on three benchmark datasets. Notably, our methodology is superior to those approaches while being significantly more efficient.
CrossNER results are not reported in the table because they are not available for competitors.

\begin{table}
    \centering
    \begin{adjustbox}{max width=\columnwidth}
    \begin{tabular}{llcGcc}
    \toprule
    \textbf{Model} & \textbf{Strategy} & \textbf{Throughput} $\uparrow$ & \textbf{GFLOPs} $\downarrow$ & \textbf{Average F1} \\
    \cmidrule(lr){1-2}\cmidrule(l{1pt}r{1pt}){3-3}\cmidrule(l{1pt}r{1pt}){4-4}\cmidrule(lr){5-5}
    \multirow{4}{*}{MiniLM}
        & Token Classif.        & \cellcolor{softgreen}\b{330.8} \small(1$\times$) & \cellcolor{softgreen}\b{1.90} \small(1$\times$) & 84.7 \\
        & PL-Marker             & \cellcolor{softorange}140.3 \small(0.42$\times$) & \cellcolor{softorange}15.8 \small(8.3$\times$) & 85.4 \\
        & \b{SpanDec (ours)}    & \cellcolor{softyellow}241.3 \small(0.73$\times$) & \cellcolor{softyellow}2.91 \small(1.5$\times$) & \b{86.2} \\
        & \b{SF-SpanDec (ours)} & \cellcolor{softgreen}310.8 \small(0.92$\times$) & \cellcolor{softgreen}1.92 \small(1.01$\times$) & 85.8 \\
                            
    \cmidrule(lr){1-2}\cmidrule(l{1pt}r{1pt}){3-3}\cmidrule(l{1pt}r{1pt}){4-4}\cmidrule(lr){5-5}
    
    \multirow{4}{*}{BERT-B}
        & Token Classif.        & \cellcolor{softgreen}\b{299.1} \small(1$\times$) & \cellcolor{softgreen}\b{7.5} \small(1$\times$) & 86.1 \\
        & PL-Marker             & \cellcolor{softorange}102.1 \small(0.34$\times$) & \cellcolor{softorange}62.5 \small(8.3$\times$) & 87.6 \\
        & \b{SpanDec (ours)}    & \cellcolor{softyellow}180.9 \small(0.61$\times$) & \cellcolor{softyellow}11.5 \small(1.5$\times$) & \b{87.8} \\
        & \b{SF-SpanDec (ours)} & \cellcolor{softgreen}276.9 \small(0.93$\times$) & \cellcolor{softgreen}7.6 \small(1.01$\times$) & 87.5 \\
    \bottomrule
    \end{tabular}
    \end{adjustbox}
 
    \caption{\b{Results for SF-SpanDec}. Throughput (samples/s), GFLOPs and average F1 scores of different strategies for MiniLM and Bert-B.}
    \label{tab:efficient}
\end{table}

\paragraph{O-Classification for Efficiency.}
In~\cref{tab:efficient}, we present the results for our efficient filtering variant, \efficientmodel{}. Compared to the standard \shortname{}, it leads to a slight performance drop; however, it still prominently outperforms the token classification baseline and remains competitive with PL-Marker. The efficiency gains of \efficientmodel{} are substantial and it delivers a strong performance boost with only a minimal increase in computational cost over standard classification. This makes \efficientmodel{} a compelling candidate for on-device and resource-constrained applications, where balancing the performance and efficiency trade-off is critical.

\vspace{-1mm}
\paragraph{Additional results.} In the appendix we report detailed results on the throughput (Sec.~\ref{sec:appendix_throughput}), the complete results for SF-SpanDec (Sec.~\ref{sec:appendix_fullresults}), an ablation study on the number of encoder/decoder blocks (Sec.~\ref{sec:appendix_ablation}) and the calculation of FLOPs (Sec.~\ref{sec:flops}).
\vspace{-0.5em}
\section{Conclusion}
\vspace{-0.3em}
\label{sec:conclusion}
In this work, we study improvement of the efficiency of span-based frameworks. More precisely, we convert an encoder only architecture to an encoder for text with a lightweight decoder in order to optimize the  utilization of span representation generation process. Additionally, deferring the span representation generation enables us to introduce a span filtering mechanism, which further reduces the computational cost of span representations. With these novel solutions, our method achieves competitive performance on NER tasks, while running fast and demonstrating a promising trade-off between accuracy and efficiency.
\vspace{-2mm}
\section{Limitations}
The hyperparameters of our method were not exhaustively optimized, leaving room for potential performance gains through further tuning. This work primarily aims to demonstrate its superiority with standard configurations in a general setting.

Furthermore, as shown in Tab.~\ref{tab:efficient}, \efficientmodel{} is so efficient that requires the same GFLOPs of naive token-classification, providing a notable boost in performance. Nevertheless, our implementation achieved 92-93\% of its throughput. This suggests possible enhancements in our software implementation, that will be addressed in future work based on the specific hardware and application. 
\clearpage

\bibliography{custom}

@inproceedings{devlin-etal-2019-bert,
    title = "{BERT}: Pre-training of Deep Bidirectional Transformers for Language Understanding",
    author = "Devlin, Jacob  and
      Chang, Ming-Wei  and
      Lee, Kenton  and
      Toutanova, Kristina",
    editor = "Burstein, Jill  and
      Doran, Christy  and
      Solorio, Thamar",
    booktitle = "Proceedings of the 2019 Conference of the North {A}merican Chapter of the Association for Computational Linguistics: Human Language Technologies, Volume 1 (Long and Short Papers)",
    month = jun,
    year = "2019",
    address = "Minneapolis, Minnesota",
    publisher = "Association for Computational Linguistics",
    url = "https://aclanthology.org/N19-1423/",
    doi = "10.18653/v1/N19-1423",
    pages = "4171--4186"
}

@article{ 10.5120/ijca2020920526,
author = { Anupama M. Nair and Anusha Aji Justus and Arjun Ramesh and Binu Rajan M. R. },
title = { Event Extraction from Emails },
journal = { International Journal of Computer Applications },
issue_date = { Jul 2020 },
volume = { 176 },
number = { 41 },
month = { Jul },
year = { 2020 },
issn = { 0975-8887 },
pages = { 1-8 },
numpages = {9},
url = { https://ijcaonline.org/archives/volume176/number41/31472-2020920526/ },
doi = { 10.5120/ijca2020920526 },
publisher = {Foundation of Computer Science (FCS), NY, USA},
address = {New York, USA}
}

@Inbook{Dozier2010,
author="Dozier, Christopher
and Kondadadi, Ravikumar
and Light, Marc
and Vachher, Arun
and Veeramachaneni, Sriharsha
and Wudali, Ramdev",
editor="Francesconi, Enrico
and Montemagni, Simonetta
and Peters, Wim
and Tiscornia, Daniela",
title="Named Entity Recognition and Resolution in Legal Text",
bookTitle="Semantic Processing of Legal Texts: Where the Language of Law Meets the Law of Language",
year="2010",
publisher="Springer Berlin Heidelberg",
address="Berlin, Heidelberg",
pages="27--43",
isbn="978-3-642-12837-0",
doi="10.1007/978-3-642-12837-0_2",
url="https://doi.org/10.1007/978-3-642-12837-0_2"
}

@Article{info10080248,
AUTHOR = {Francis, Sumam and Van Landeghem, Jordy and Moens, Marie-Francine},
TITLE = {Transfer Learning for Named Entity Recognition in Financial and Biomedical Documents},
JOURNAL = {Information},
VOLUME = {10},
YEAR = {2019},
NUMBER = {8},
ARTICLE-NUMBER = {248},
URL = {https://www.mdpi.com/2078-2489/10/8/248},
ISSN = {2078-2489},
DOI = {10.3390/info10080248}
}

@article{liu2020crossner,
      title={CrossNER: Evaluating Cross-Domain Named Entity Recognition}, 
      author={Zihan Liu and Yan Xu and Tiezheng Yu and Wenliang Dai and Ziwei Ji and Samuel Cahyawijaya and Andrea Madotto and Pascale Fung},
      year={2020},
      eprint={2012.04373},
      archivePrefix={arXiv},
      primaryClass={cs.CL}
}

@inproceedings{bogdanov-etal-2024-nuner,
    title = "{N}u{NER}: Entity Recognition Encoder Pre-training via {LLM}-Annotated Data",
    author = "Bogdanov, Sergei  and
      Constantin, Alexandre  and
      Bernard, Timoth{\'e}e  and
      Crabb{\'e}, Benoit  and
      Bernard, Etienne P",
    editor = "Al-Onaizan, Yaser  and
      Bansal, Mohit  and
      Chen, Yun-Nung",
    booktitle = "Proceedings of the 2024 Conference on Empirical Methods in Natural Language Processing",
    month = nov,
    year = "2024",
    address = "Miami, Florida, USA",
    publisher = "Association for Computational Linguistics",
    url = "https://aclanthology.org/2024.emnlp-main.660/",
    doi = "10.18653/v1/2024.emnlp-main.660",
    pages = "11829--11841"
}

@inproceedings{zhong-chen-2021-frustratingly,
    title = "A Frustratingly Easy Approach for Entity and Relation Extraction",
    author = "Zhong, Zexuan  and
      Chen, Danqi",
    editor = "Toutanova, Kristina  and
      Rumshisky, Anna  and
      Zettlemoyer, Luke  and
      Hakkani-Tur, Dilek  and
      Beltagy, Iz  and
      Bethard, Steven  and
      Cotterell, Ryan  and
      Chakraborty, Tanmoy  and
      Zhou, Yichao",
    booktitle = "Proceedings of the 2021 Conference of the North American Chapter of the Association for Computational Linguistics: Human Language Technologies",
    month = jun,
    year = "2021",
    address = "Online",
    publisher = "Association for Computational Linguistics",
    url = "https://aclanthology.org/2021.naacl-main.5/",
    doi = "10.18653/v1/2021.naacl-main.5",
    pages = "50--61"
}

@inproceedings{wang-etal-2021-automated,
    title = "Automated Concatenation of Embeddings for Structured Prediction",
    author = "Wang, Xinyu  and
      Jiang, Yong  and
      Bach, Nguyen  and
      Wang, Tao  and
      Huang, Zhongqiang  and
      Huang, Fei  and
      Tu, Kewei",
    editor = "Zong, Chengqing  and
      Xia, Fei  and
      Li, Wenjie  and
      Navigli, Roberto",
    booktitle = "Proceedings of the 59th Annual Meeting of the Association for Computational Linguistics and the 11th International Joint Conference on Natural Language Processing (Volume 1: Long Papers)",
    month = aug,
    year = "2021",
    address = "Online",
    publisher = "Association for Computational Linguistics",
    url = "https://aclanthology.org/2021.acl-long.206/",
    doi = "10.18653/v1/2021.acl-long.206",
    pages = "2643--2660"
}

@inproceedings{ye-etal-2022-packed,
    title = "Packed Levitated Marker for Entity and Relation Extraction",
    author = "Ye, Deming  and
      Lin, Yankai  and
      Li, Peng  and
      Sun, Maosong",
    editor = "Muresan, Smaranda  and
      Nakov, Preslav  and
      Villavicencio, Aline",
    booktitle = "Proceedings of the 60th Annual Meeting of the Association for Computational Linguistics (Volume 1: Long Papers)",
    month = may,
    year = "2022",
    address = "Dublin, Ireland",
    publisher = "Association for Computational Linguistics",
    url = "https://aclanthology.org/2022.acl-long.337/",
    doi = "10.18653/v1/2022.acl-long.337",
    pages = "4904--4917"
}

@misc{seqeval,
  title={{seqeval}: A Python framework for sequence labeling evaluation},
  url={https://github.com/chakki-works/seqeval},
  note={Software available from https://github.com/chakki-works/seqeval},
  author={Hiroki Nakayama},
  year={2018},
}

@misc{Aarsen_SpanMarker,
  author = {Tom Aarsen},
  title = {SpanMarker for Named Entity Recognition},
  year = {2023},
  publisher = {GitHub},
  journal = {GitHub repository},
  howpublished = {\url{https://github.com/tomaarsen/SpanMarkerNER}},
}

@article{zhou2023universalner,
  title={Universalner: Targeted distillation from large language models for open named entity recognition},
  author={Zhou, Wenxuan and Zhang, Sheng and Gu, Yu and Chen, Muhao and Poon, Hoifung},
  journal={arXiv preprint arXiv:2308.03279},
  year={2023}
}

@inproceedings{zhu2024open,
  title={Open-source large language models excel in named entity recognition},
  author={Zhu, Dengya and Li, Sirui and Thompson, Nik and Wong, Kok Wai},
  booktitle={International Conference on Neural Information Processing},
  pages={313--326},
  year={2024},
  organization={Springer}
}

@inproceedings{zaratiana-etal-2024-gliner,
  title     = {{GL}i{NER}: Generalist Model for Named Entity Recognition using Bidirectional Transformer},
  author    = {Zaratiana, Urchade and others},
  booktitle = {Proceedings of the 2024 Conference of the North American Chapter of the Association for Computational Linguistics (NAACL)},
  year      = {2024},
  url       = {https://aclanthology.org/2024.naacl-long.300/}
}

@inproceedings{arora-park-2023-split-ner,
  title     = {Split-{NER}: Named Entity Recognition via Two Question-Answering-based Classifications},
  author    = {Arora, Jatin and Park, Youngja},
  booktitle = {Proceedings of the 61st Annual Meeting of the Association for Computational Linguistics (Volume 2: Short Papers)},
  year      = {2023},
  url       = {https://aclanthology.org/2023.acl-short.36/}
}

@inproceedings{kim-etal-2024-verifiner,
  title     = {Verifi{NER}: Verification-augmented {NER} via Knowledge-grounded Reasoning with Large Language Models},
  author    = {Kim, Seoyeon and Seo, Kwangwook and Chae, Hyungjoo and Yeo, Jinyoung and Lee, Dongha},
  booktitle = {Proceedings of the 62nd Annual Meeting of the Association for Computational Linguistics (Volume 1: Long Papers)},
  year      = {2024},
  url       = {https://aclanthology.org/2024.acl-long.134/}
}

@inproceedings{wadden-etal-2019-dygiepp,
  title     = {Entity, Relation, and Event Extraction with Contextualized Span Representations},
  author    = {Wadden, David and Wennberg, Ulme and Luan, Yi and Hajishirzi, Hannaneh},
  booktitle = {Proceedings of the 2019 Conference on Empirical Methods in Natural Language Processing and the 9th International Joint Conference on Natural Language Processing (EMNLP-IJCNLP)},
  year      = {2019},
  url       = {https://aclanthology.org/D19-1585/}
}

@inproceedings{wang-etal-2025-gpt,
  title = "{GPT}-{NER}: Named Entity Recognition via Large Language Models",
  author = "Wang, Shuhe  and
    Sun, Xiaofei  and
    Li, Xiaoya  and
    Ouyang, Rongbin  and
    Wu, Fei  and
    Zhang, Tianwei  and
    Li, Jiwei  and
    Wang, Guoyin  and
    Guo, Chen",
  booktitle = "Findings of the Association for Computational Linguistics: NAACL 2025",
  year = "2025",
  month = apr,
  address = "Albuquerque, New Mexico",
  publisher = "Association for Computational Linguistics",
  url = "https://aclanthology.org/2025.findings-naacl.239/",
  doi = "10.18653/v1/2025.findings-naacl.239",
  pages = "4257--4275"
}

@inproceedings{wang-etal-2020-minilm,
  title     = {MiniLM: Deep Self-Attention Distillation for Task-Agnostic Compression of Pre-Trained Transformers},
  author    = {Wang, Wenhui and Wei, Furu and Dong, Li and Bao, Hangbo and Yang, Nan and Zhou, Ming},
  booktitle = {Advances in Neural Information Processing Systems (NeurIPS)},
  year      = {2020},
  url       = {https://arxiv.org/abs/2002.10957}
}

@misc{wang-etal-2023-instructuie,
  title        = {Multi-task Instruction Tuning for Unified Information Extraction},
  author       = {Wang, Xingyao and Zhou, Wenxuan and Zu, Chao and Xia, Haoran and Chen, Tian and Zhang, Yong and Zheng, Rui and Ye, Jing and Zhang, Qianying and Gui, Tao and Kang, Jian and Yang, Junfeng and Li, Sha and Du, Chao},
  year         = {2023},
  eprint       = {2304.08085},
  archivePrefix= {arXiv},
  primaryClass = {cs.CL},
  url          = {https://arxiv.org/abs/2304.08085}
}

@inproceedings{wolf-etal-2020-transformers,
    title = "Transformers: State-of-the-Art Natural Language Processing",
    author = "Wolf, Thomas and Debut, Lysandre and Sanh, Victor and others",
    booktitle = "Proceedings of the 2020 Conference on Empirical Methods in Natural Language Processing: System Demonstrations",
    year = "2020",
    publisher = "Association for Computational Linguistics",
    url = "https://www.aclweb.org/anthology/2020.emnlp-demos.6",
    pages = "38--45"
}

@inproceedings{hovy-etal-2006-ontonotes,
    title = "{O}nto{N}otes: The 90{\%} Solution",
    author = "Hovy, Eduard  and
      Marcus, Mitchell  and
      Palmer, Martha  and
      Ramshaw, Lance  and
      Weischedel, Ralph",
    booktitle = "Proceedings of the Human Language Technology Conference of the {NAACL}, Companion Volume: Short Papers",
    month = jun,
    year = "2006",
    address = "New York City, USA",
    publisher = "Association for Computational Linguistics",
    url = "https://aclanthology.org/N06-2015",
    pages = "57--60",
}

@article{DBLP:journals/biodb/LiSJSWLDMWL16,
  author    = {Jiao Li and
               Yueping Sun and
               Robin J. Johnson and
               Daniela Sciaky and
               Chih{-}Hsuan Wei and
               Robert Leaman and
               Allan Peter Davis and
               Carolyn J. Mattingly and
               Thomas C. Wiegers and
               Zhiyong Lu},
  title     = {BioCreative {V} {CDR} task corpus: a resource for chemical disease
               relation extraction},
  journal   = {Database J. Biol. Databases Curation},
  volume    = {2016},
  year      = {2016},
  url       = {https://doi.org/10.1093/database/baw068},
  doi       = {10.1093/database/baw068},
  bibsource = {dblp computer science bibliography, https://dblp.org}
}

@inproceedings{wang2019crossweigh,
  title={CrossWeigh: Training Named Entity Tagger from Imperfect Annotations},
  author={Wang, Zihan and Shang, Jingbo and Liu, Liyuan and Lu, Lihao and Liu, Jiacheng and Han, Jiawei},
  booktitle={Proceedings of the 2019 Conference on Empirical Methods in Natural Language Processing and the 9th International Joint Conference on Natural Language Processing (EMNLP-IJCNLP)},
  pages={5157--5166},
  year={2019}
}

@inproceedings{tjong-kim-sang-de-meulder-2003-introduction,
    title = "Introduction to the {C}o{NLL}-2003 Shared Task: Language-Independent Named Entity Recognition",
    author = "Tjong, Kim Sang and Erik, F. and 
      De Meulder, Fien",
    booktitle = "Proceedings of the Seventh Conference on Natural Language Learning at {HLT}-{NAACL} 2003",
    year = "2003",
    url = "https://www.aclweb.org/anthology/W03-0419",
    pages = "142--147",
}

\clearpage
\appendix
\section{Throughput}
\label{sec:appendix_throughput}
We evaluate the throughput (samples/s) for each method on all the considered dataset and we report these numbers in Table~\ref{tab:Throughput}. To provide comparable results, we use the same setup and inference parameters for all the models (single Nvidia A40 GPU, mixed precision, etc.). The results are the average of 3 runs.

\section{Full Results for SF-SpanDec}
\label{sec:appendix_fullresults}
Additionally, in Table~\ref{tab:full_results_efficient}, we report the full results of our efficient model (SF-SpanDec) that are previously reported average results in Tab.~\ref{tab:efficient}, compared to our standard model (SpanDec), PL-Marker and token-classification baseline

\begin{table*}[h!]
    \centering
    \begin{adjustbox}{max width=0.9\textwidth}
    \begin{tabular}{llccccc }
    \toprule
    \textbf{Model} & \textbf{Strategy} & \textbf{CoNLL++} & \textbf{CrossNER} & \textbf{OntoNotes5} & \textbf{BC5CDR} & \textbf{Average}\\
    \cmidrule(lr){1-2}\cmidrule(lr){3-6}\cmidrule(lr){7-7}
    \multirow{3}{*}{MiniLM} & Token Classif.      & 247.6 & 297.3 & 350.0 & 428.1 & 330.8 \\
                            & PL-Marker           & 86.0  & 109.1 & 123.6 & 242.6 & 140.3 \\
                            & \b{SpanDec (ours)}  & 185.7 & 221.0 & 243.2  & 315.3 & 241.3 \\
                            & \b{SF-SpanDec (ours)}  & 258.5 & 266.6 & 319.1  & 399.0 & 310.8 \\
                            
    \cmidrule(lr){1-2}\cmidrule(lr){3-6}\cmidrule(lr){7-7}
    
    \multirow{3}{*}{BERT-B}   & Token Classif.  & 206.3 & 240.5 & 330.9 & 418.6 & 299.1\\
    & PL-Marker                                 & 64.1  & 78.1  & 84.3 & 182.0 & 102.1 \\
    & \textbf{SpanDec (ours)}                   & 114.3 & 163.8 & 161.4 & 283.9 & 180.9 \\
    & \b{SF-SpanDec (ours)}                     & 230.1 & 212.2 & 282.0  & 383.3 & 276.9 \\
    \cmidrule(lr){1-2}\cmidrule(lr){3-6}\cmidrule(lr){7-7}
    
    \multirow{3}{*}{RoBERTa-L}  & Token Classif.    & 131.4 & 127.0 & 221.2 & 268.5 & 187.0 \\
    & PL-Marker                                     & 40.2  & 59.5 & 60.8 & 111.5 & 68.0 \\
    & \textbf{SpanDec (ours)}                       & 71.2  & 92.9 & 110.4 & 200.4 & 118.7 \\

    \bottomrule
    \end{tabular}
    \end{adjustbox}
    \caption{\b{Throughput Results.} Throughput (samples/s) evaluated using a single Nvidia-A40 GPU with a reference batch size of 8.}
    \label{tab:Throughput}
\end{table*}
\begin{table*}[h!]
    \centering
    \begin{adjustbox}{max width=0.9\textwidth}
    \begin{tabular}{llccccc}
    \toprule
    \textbf{Model} & \textbf{Strategy} & \textbf{CoNLL++} & \textbf{CrossNER} & \textbf{OntoNotes5} & \textbf{BC5CDR} & \textbf{Average f1} \\
    \cmidrule(lr){1-2}\cmidrule(lr){3-6}\cmidrule(lr){7-7}
    \multirow{3}{*}{MiniLM} & Token Classif.          & 92.9     & 71.2     & 87.8     & 87.1 & 84.7       \\
                            & PL-Marker               & 91.7     & 73.9     & 87.9     & \b{88.0} & 85.4    \\
                            & \b{SpanDec (ours)}      & \b{93.7} & \b{74.6} & \b{88.4} & \b{88.0} & \b{86.2} \\
                            & \textbf{SF-SpanDec (ours)}  & 93.4 & 73.8 & 88.3 & 87.6 & 85.8 \\
    \cmidrule(lr){1-2}\cmidrule(lr){3-6}\cmidrule(lr){7-7}
    
    \multirow{3}{*}{BERT-B}      & Token Classif.  & 92.9 & 74.5 & 89.5 & 87.4 & 86.1     \\
    & PL-Marker                  & 93.2 & 78.5 & 90.0 & \b{88.5} & 87.6                 \\
    & \textbf{SpanDec (ours)}    & \b{93.6} & \b{78.7} & \b{90.3} & 88.4 & \b{87.8}      \\
    & \textbf{SF-SpanDec (ours)} & 93.4 & 78.2 & 90.1 & 88.2 & 87.5 \\

    \bottomrule
    \end{tabular}
    \end{adjustbox}
    \caption{\b{Full Results for SF-SpanDec.} F1 score on CoNLL++, CrossNER, OntoNotes5 and BC5CDR benchmarks using different encoder models. The best performance per dataset and encoder model is highlighted in \textbf{bold}.}
    \label{tab:full_results_efficient}
\end{table*}

\section{Ablation Study: Number of Encoder and Decoder Blocks}
\label{sec:appendix_ablation}

Table~\ref{tab:ablations} presents an ablation study investigating the impact of varying the number of encoder and decoder layers within our SpanDec model architecture. This study is conducted using the MiniLM model on the CoNLL++ dataset. Two sets of experiments are performed:

\begin{enumerate}
    \item \textbf{Encoder Reduction, Fixed Decoder:} In the first set, we progressively remove layers from the encoder, starting from the final layer, while maintaining a fixed single-layer decoder.
    \item \textbf{Constant Total Layers:} In the second set, for each layer remove from the encoder, a corresponding layer is added to the decoder. This maintains a constant total number of 12 encoder and decoder layers.
\end{enumerate}

The results from the first experiment indicate that reducing the number of encoder layers leads to a modest decrease in performance. Crucially, this reduction also corresponds to a decrease in computational cost, measured in GFLOPs, suggesting this configuration could be beneficial for applications with computational constraints.

Conversely, the second experiment demonstrates that increasing the number of decoder layers does not yield significant performance improvements while simultaneously increasing computational complexity. This suggests that a simple model, single-layer decoder, is sufficient for the Named Entity Recognition task with this architecture.

\begin{table}[h!]

    \centering
    \centering
    \begin{adjustbox}{max width=\columnwidth}
    \begin{tabular}{cccc}
    
    \toprule
    \textbf{Encoder Blocks} & \textbf{Decoder Blocks} & \textbf{GFLOPs} & \textbf{F1} \\
    \toprule
    11 (31.5M) & 1 (1.8M) & 2.9 & 93.6 \\
    \midrule
    \midrule
    10 (29.7M) & 1 (1.8M) & 2.8 & 93.4 \\
    9  (27.9M) & 1 (1.8M) & 2.6 & 93.2 \\
    8  (26.1M) & 1 (1.8M) & 2.4 & 92.9 \\
    7  (24.3M) & 1 (1.8M) & 2.3 & 92.3 \\ 
    \midrule
    \midrule
    10 (29.7M) & 2 (1.8M) & 3.9 & 93.4 \\
    9  (27.9M) & 3 (3.6M) & 4.9 & 93.2 \\
    8  (26.1M) & 4 (5.4M) & 5.9 & 93.2 \\
    7  (24.3M) & 5 (9.0M) & 7.0 & 92.4 \\ 
    
    \bottomrule
    \end{tabular}
    \end{adjustbox}
    
    \caption{\b{Ablation study: Impact of Encoder/Decoder Layers.} In the top section, we report the base configuration for SpanDec ($11$ encoder layers and $1$ encoder layer). In the middle section, we remove the final layers of the encoder. In the bottom section, we keep constant the total number of layers to $12$.}
    \label{tab:ablations}
\end{table}
\section{FLOPs Formulas for Transformer Encoder}
\label{sec:flops}
Profiling tool, such as \texttt{torch.profiler}, can be inaccurate in estimating the floating-point operations (FLOPs) of the attention mechanism, e.g.\ when fused kernels are used to optimize the computation device performance. Therefore, we conduct the following analytic approach to estimate the FLOPs of different models and approaches adopted in this work.

Since the computation of the input embedding functions, the output linear classifier, activation functions and normalization layers are negligible (in total $\lesssim 2\%$) compared to the computation of the attention mechanism and the FFN (feed-forward network) linear projection, our analysis will focus on the last two operations. 

Parameters of the calculation formulas are defined as follows:
\begin{itemize}
    \item $L$: Number of Transformer layers.
    \item $S$: Input sequence length.
    \item $M$: Span marker lengths, $M$ indicates $M/2$ pairs of span markers.
    \item $H_{dim}$: Hidden dimension size.
    \item $A$: Number of attention heads.
    \item $H_{head} = H_{dim} / A$: Dimension size per attention head.
    \item $F_{dim}$: FFN intermediate dimension size (typically $4 \times H_{dim}$)
\end{itemize}
Given an architecture, these parameters are fixed except $S$ and $M$. We summarize the parameters corresponding to the models adopted in our experiments in \cref{tab:model_params}.

\begin{table}[t!]
\centering

\resizebox{\columnwidth}{!}{
\begin{tabular}{@{}lccc@{}}
\toprule
\textbf{Parameter}                     & \textbf{MiniLM} & \textbf{BERT-B} & \textbf{RoBERTa-L} \\
\midrule
$L$ (Transformer blocks)    & 12                                         & 12                         & 24                     \\
$H_{dim}$ (Hidden size)   & 384                                        & 768                        & 1024                   \\
$A$ (Attention heads)       & 12                                         & 12                         & 16                     \\
$F_{dim}$ (FFN int. size) & $1536$                      & $3072$      & $4096$ \\
\bottomrule
\end{tabular}
}
\caption{\b{Architecture Details.} Configuration details for different models.}
\label{tab:model_params}
\end{table}
Next we first present the FLOPs calculation for the text tokens when bidirectional self-attention is applied. Note that 1 MAC (multiply-accumulate) operation is generally counted as 2 FLOPs.

\paragraph{\textbf{1. Bidirectional Self-Attention for Text Tokens} ($Attn_{text}$ per block).}
The self-attention mechanism of text tokens mainly consists of:
\begin{itemize}
    \item Input/Output Projections (for Q, K, V, and the final output projection; 4 matrices of $H_{dim} \times H_{dim}$ applied to $S$ tokens): $4 \times (2 \cdot S \cdot H_{dim} \cdot H_{dim}) = 8 S H_{dim}^2$.
    \item Attention Score Calculation ($Q K^T$ across $A$ heads): $A \cdot (2 \cdot S \cdot H_{head} \cdot S) = 2 S^2 (A \cdot H_{head}) = 2 S^2 H_{dim}$.
    \item Weighted Sum with V (applying attention scores to V, across $A$ heads): $A \cdot (2 \cdot S \cdot H_{head} \cdot S) = 2 S^2 (A \cdot H_{head}) = 2 S^2 H_{dim}$.
\end{itemize}
The total FLOPs for self-attention per block is:
\begin{equation}
    \label{eq:attn_text}
    Attn_{text} = 8 S H_{dim}^2 + 4 S^2 H_{dim}
\end{equation}

\paragraph{\textbf{2. FFN Linear Projection for Text Tokens} ($FFN_{text}$ per block)}
Bert-like model's FFN typicall includes two linear layers:
\begin{itemize}
    \item First Linear Layer (expansion: $H_{dim} \rightarrow F_{dim}$): $2 \cdot S \cdot H_{dim} \cdot F_{dim}$.
    \item Second Linear Layer (contraction: $F_{dim} \rightarrow H_{dim}$): $2 \cdot S \cdot F_{dim} \cdot H_{dim}$.
\end{itemize}
The total FLOPs for the FFN per block is:
\begin{equation}
    \label{eq:ffn_text}
    FFN_{text} = 4 S H_{dim} F_{dim}
\end{equation}

Then, we present the FLOPs calculation for the span markers. In particular, queries of a pair of marker tokens matches the keys of each other (omitted in this analysis for simplicity) plus the keys of input text tokens, resembling the cross-attention mechanism. Note that the computation of span tokens is additional to the previous computation of text tokens. It consists of the following:

\paragraph{\textbf{Cross-Attention for Span Markers} ($Attn_{marker}$ per block)}
The cross-attention mechanism mainly consists of:
\begin{itemize}
    \item Input/Output Projections (for Q, K, V, and the final output projection; 4 matrices of $H_{dim} \times H_{dim}$ applied to $M$ tokens): $4 \times (2 \cdot M \cdot H_{dim} \cdot H_{dim}) = 8 M H_{dim}^2$.
    \item Attention Score Calculation ($Q K^T$ across $A$ heads): $A \cdot (2 \cdot S \cdot H_{head} \cdot M) = 2 SM (A \cdot H_{head}) = 2 SM H_{dim}$.
    \item Weighted Sum with V (applying attention scores to V, across $A$ heads): $A \cdot (2 \cdot S \cdot H_{head} \cdot M) = 2 SM (A \cdot H_{head}) = 2 SM H_{dim}$.
\end{itemize}
The total FLOPs for cross-attention per block is:
\begin{equation}
    \label{eq:attn_marker}
    Attn_{marker} = 8 M H_{dim}^2 + 4 SM H_{dim}
\end{equation}

\paragraph{\textbf{FFN Span Markers} ($FFN_{marker}$ per block)}
Following the calculation of FFN's cost for text tokens, we can derive the total FLOPs per block is:
\begin{equation}
    \label{eq:ffn_marker}
    FFN_{marker} = 4 M H_{dim}F_{dim}
\end{equation}

Given the parameters in Tab.~\ref{tab:model_params}, the number of layers involving span marker computation, and the lengths of input text and span marker, we can calculate the FLOPs based on \cref{eq:attn_text,eq:ffn_text,eq:attn_marker,eq:ffn_marker}. Based on the average input length of the datasets adopted in the work, we set $S=44$ and $M=324$ for the FLOPs computation.

\end{document}